\documentclass[letterpaper, 10pt, conference]{ieeeconf}  

\IEEEoverridecommandlockouts                              
                                                          
\usepackage{gensymb}
\usepackage{times}
\usepackage{epsfig}
\usepackage{graphicx}
\usepackage{amsmath}
\usepackage{amsfonts}
\usepackage{amssymb}
\usepackage{bm}
\usepackage{comment}
\usepackage{longtable}
\usepackage{hhline}
\usepackage{ctable} 
\usepackage{multirow}
\usepackage{cite}
\usepackage{xcolor}
\usepackage{subcaption}
\usepackage{url}
\usepackage{subfig}
\usepackage{makecell}

\DeclareUrlCommand\url{\color{blue}}

\title{\LARGE \bf Endangered Alert: A Field-Validated Self-Training Scheme for Detecting and Protecting Threatened Wildlife on Roads and Roadsides}

\author{
    Kunming Li$^{1}$, Mao Shan$^{1*}$, Stephany Berrio Perez$^{1}$, Katie Luo$^{2}$, and Stewart Worrall$^{1}$
    \thanks{$^1$ The authors are with the Australian Centre for Robotics (ACFR) at The University of Sydney (NSW, Australia). E-mails: {\tt\small{\{k.li, m.shan, j.berrio, s.worrall\}@acfr.usyd.edu.au}}}
    \thanks{$^2$ This author is with the Computer Information Sciences Department at Cornell University (NY, USA). E-mail:
        {\tt\small kzl6@cornell.edu}}
    \thanks{$^*$ Corresponding author. E-mail: {\tt\small m.shan@acfr.usyd.edu.au}}
    \thanks{$^\dag$ This research is funded by the Queensland Government's Department of Transport and Main Roads, and iMOVE CRC and supported by the Cooperative Research Centres program, an Australian Government initiative.}
    \thanks{$^\ddag$ This study was approved by the University of Sydney Animal Ethics Committee (project number: 2023/2398).}
}

\begin{document}

\maketitle
\thispagestyle{empty}
\pagestyle{empty}

\begin{abstract}
Traffic accidents are a global safety concern, resulting in numerous fatalities each year. A considerable number of these deaths are caused by animal-vehicle collisions (AVCs), which not only endanger human lives but also present serious risks to animal populations.
This paper presents an innovative self-training methodology aimed at detecting rare animals, such as the cassowary in Australia, whose survival is threatened by road accidents.
The proposed method addresses critical real-world challenges, including acquiring and labelling sensor data for rare animal species in resource-limited environments. It achieves this by leveraging cloud and edge computing, and automatic data labelling to improve the detection performance of the field-deployed model iteratively. 
Our approach introduces Label-Augmentation Non-Maximum Suppression (LA-NMS), which incorporates a vision-language model (VLM) to enable automated data labelling. During a five-month deployment, we confirmed the method’s robustness and effectiveness, resulting in improved object detection accuracy and increased prediction confidence. The source code is available: \url{https://github.com/acfr/CassDetect}.
\end{abstract}

\section{INTRODUCTION}

According to a recent report from the World Health Organisation (WHO) \cite{world2019global}, road safety remains a global concern, with nearly 1.19 million people dying annually and another 20 to 50 million sustaining non-fatal injuries due to traffic accidents. Some of these accidents involve animal-vehicle collisions (AVCs), which not only threaten human lives but also lead to wildlife fatalities. In Australia, it is estimated that 10 million animals die each year in road incidents \cite{IAG_Wildlife_Road_Safety}. The situation is particularly dire for endangered species such as the cassowary, whose road mortality rates increase their risk of extinction.

Geographical and financial limitations often restrict traditional mitigation strategies like wildlife crossings and fencing. In response, there has been a growing interest in developing Roadside Animal Detection Systems (RADS) and similar technologies to mitigate conflicts between animals and road networks, yet they come with challenges. Some RADS can disrupt animal behaviours with artificial stimuli such as lights and sounds \cite{navtech,sharafsaleh2012evaluation}. Many are hindered by high costs for installation and maintenance \cite{druta2015evaluation, druta2020preventing}, limiting their widespread adoption. Additionally, questions persist about the scalability and adaptability of these technologies across diverse geographic and climatic conditions \cite{munian2022intelligent}.

Recent advancements in machine learning (ML) offer promising avenues for enhancing RADS. However, these approaches face challenges, particularly for underrepresented species in available datasets.

One of the challenges of training ML models for animal detection lies in acquiring and labelling the required amount of data. Collecting sensor data for rare animal species, such as cassowaries, presents considerable logistical and financial hurdles. For instance, during the field trial, we recorded over 60,000 vehicles but observed only 17 cassowaries on average per week. Relying on traditional data labelling requires extensive human effort, which adds to these difficulties.

In addition, RADS deployed in remote areas face resource limitations, including restricted data transfer capabilities and limited computing power. This limits the amount of data stored and the frequent updating of ML models, which is important for maintaining system efficacy.

\begin{figure}[!t]
	\centering
	\begin{subfigure}[]{0.88\columnwidth}
        \label{fig:system:a}
        \includegraphics[width=\columnwidth]{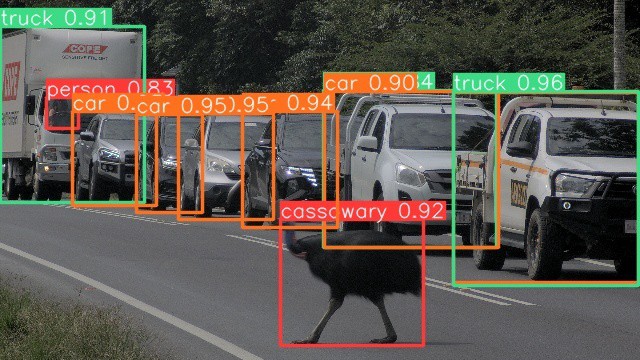}
        \caption{Cassowary and road users detection}
    \end{subfigure}
    \begin{subfigure}[]{0.88\columnwidth}
        \label{fig:system:b}
        \includegraphics[width=\columnwidth]{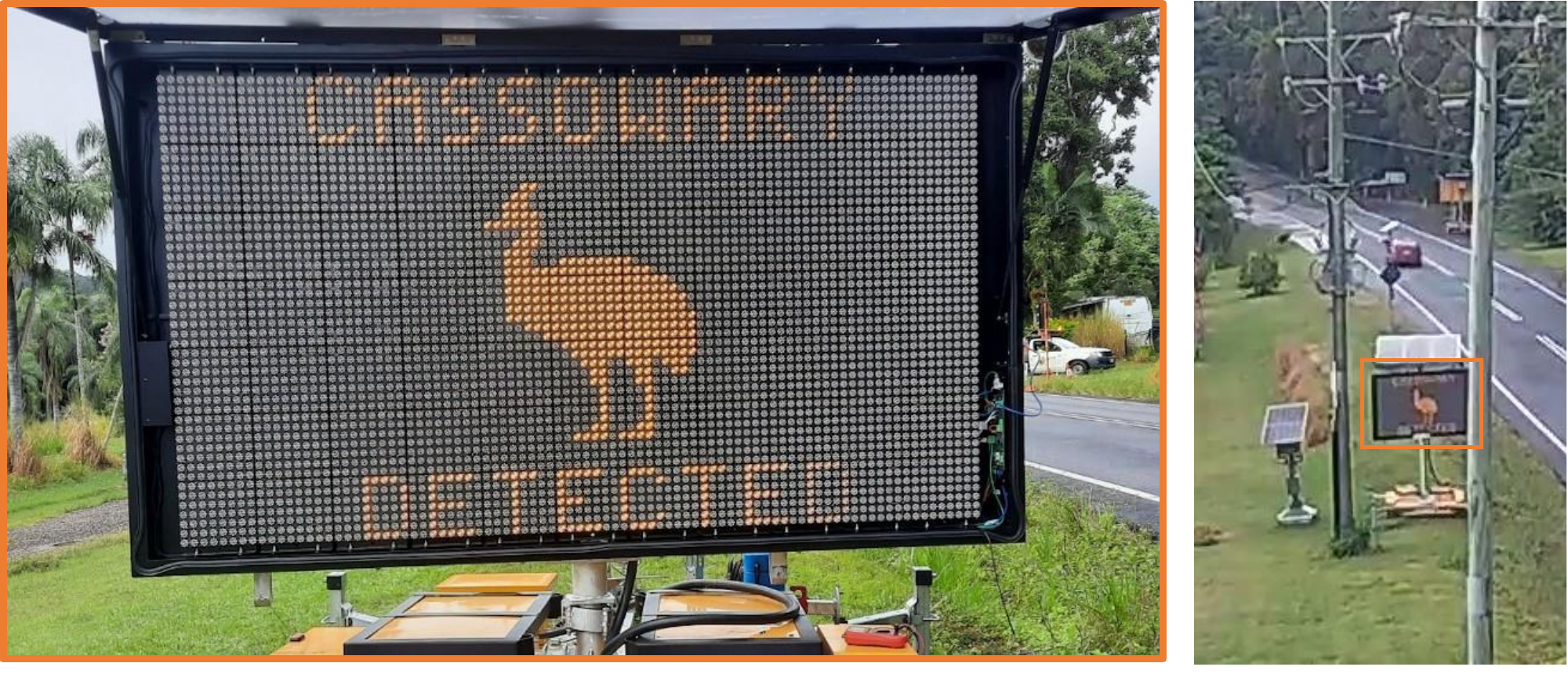}
        \caption{Variable message sign}
    \end{subfigure}
    \caption{\small Deployed system identifying a cassowary crossing a busy road. Upon detecting a cassowary, as illustrated in (a), the system alerts approaching vehicles through a variable message sign (VMS) to help prevent potential collisions, as shown in (b). (Message design credit: Ioni Lewis, Queensland University of Technology)}
	\label{fig:system}
\end{figure}

This paper introduces a novel and adaptive self-training methodology for roadside units aiming to detect rare animals in remote locations. The contribution of this paper addresses: 
\begin{enumerate}
    \item Self-training pipeline in resource-constrained environments.
    \item Automatic data labelling using Label-Augmentation Non-Maximum Suppression (LA-NMS) for vision-language models (VLMs).
    \item A five-month field validation to assess the effectiveness of the proposed method.
\end{enumerate}

The developed system is intended to detect cassowaries on and near the road to alert drivers and thus avoid potential AVCs, as presented in \textit{Fig. \ref{fig:system}}. The five-month long field trial was conducted from February to June 2024 in Far North Queensland, Australia. 

\section{Related Work}

\subsection{Label Efficient Learning for Object Detection}
Recent advances in robotics, natural language processing (NLP), and computer vision have enabled their deployment in diverse environments where an abundance of manually labelled data is readily available to train existing deep learning algorithms. However, manual data annotation is financially prohibitive at large scales. It limits the deployment of such algorithms in underrepresented environments like mines, burning buildings, and wilderness areas, where labelled data is not readily available. To address this, various robotics paradigms have been proposed to train models with limited supervision, including self-supervised learning (SSL), semi-supervised learning object detection (SSOD), and weakly supervised learning object detection (WSOD)\cite{tang2017multiple, lin2020object, ren2020instance, tang2018pcl, yin2021instance, huang2020comprehensive,cao2021cat,dong2021boosting,zhong2020boosting, bilen2016weakly}. \cite{tang2017multiple, ren2020instance} improve WSOD by utilising psudo groundtruth mining. Authors in \cite{bilen2016weakly,wan2019continuation} improve WSOD by integrating multi-instance learning (MIL) framework. SSOD trains object detectors using a limited set of images with precise annotations alongside a substantial collection of unlabelled images \cite{wang2018towards, jeong2019consistency, sohn2020simple, liu2021unbiased, yang2021interactive, misra2015watch, tang2016large, roychowdhury2019automatic, rosenberg2005semi}. Among the approaches developed for SSOD, SSM \cite{wang2018towards} incorporates high-confidence patches from unlabelled images as pseudo labels to enhance training. \cite{jeong2019consistency} focuses on improving data consistency and eliminating background distractions, while STAC \cite{sohn2020simple} leverages extensive data augmentation techniques on unlabelled images to enrich model robustness. Additionally, \cite{liu2021unbiased} implements a teacher-student framework that utilises knowledge distillation to improve the learning process in SSOD. Similarly, \cite{yang2021interactive} adopts a mean teacher strategy, wherein a more stable and consistent model guides the learning of the primary model. These methods strive to optimise limited labelled data and maximise learning from unlabelled datasets. However, they still face challenges, such as the requirement for noise-free annotations and a balanced split of labelled and unlabelled data, which are not always achievable in practical scenarios. Furthermore, the use of rich feature representations generated by emergent vision foundation models like DINO \cite{ren2024grounding, liu2023grounding}, CLIP \cite{radford2021learning}, SAM \cite{kirillov2023segment} and OWL \cite{minderer2024scaling} reduces or eliminates the need for manual data annotation in existing training protocols.

\subsection{Machine Learning Models on Field Robots and Edge Device}
Self-training methods \cite{liu2022self, sudharsan2021train++, wang2021list} typically involve training a model on the device with labelled data and then making predictions on unlabelled data. If the top prediction score for an unlabelled input exceeds a threshold, the input is pseudo-labelled and used in further training iterations. While this approach can enhance performance, it also slows down training and can lead to instability depending on the threshold. Additionally, many edge devices and field robots cannot access labelled data or may have noisy labels for initial model training, hindering the effectiveness of self-training using top prediction scores.

\begin{figure*}[!t]
\centering
\includegraphics[width=0.9\textwidth]{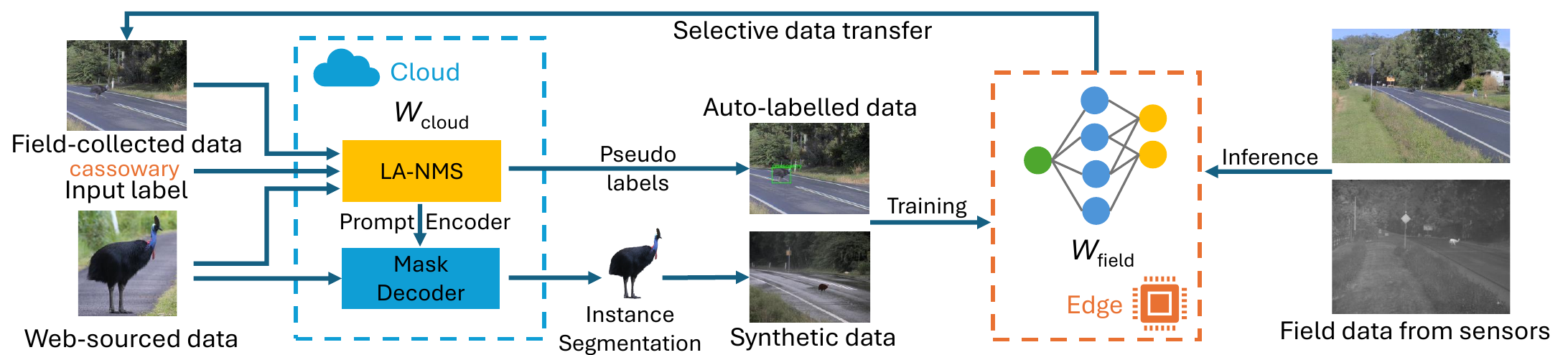}
\caption{\small Overview of the proposed self-training ML scheme for roadside animal detection. Initially, the cloud-based model \(W_{\text{cloud}}\) synthesises images of cassowaries and generates the pseudo-labels using the web-sourced cassowary images and field background images. These images are used to train the initial field detection model for deployment on the edge device. In the deployment environment, this field model \(W_{\text{field}}\) processes images captured in real time and selects relevant data to send back to the cloud server. \(W_{\text{cloud}}\) then automatically processes the received field data, generating pseudo-labels, which are used to fine-tune \(W_{\text{field}}\). This iterative cycle progressively improves the detection performance of \(W_{\text{field}}\). Note that the pseudo-labelling task is performed using the proposed LA-NMS approach, which integrates a VLM (i.e., OWL-VIT \cite{minderer2024scaling} in our work), within \( W_{\text{cloud}} \).}
\label{fig:stage_overview}
\end{figure*}

Another approach involves adapting and fine-tuning pre-trained ML models directly at the edge, eliminating the need for complete retraining. Various studies \cite{cai2020tinytl, yang2022rep,chiang2023mobiletl, yang2021joint, choi2022accelerating} propose transfer learning for edge learning, allowing models on edge devices to adapt and fine-tune with minimal computational resources. Online adaptive learning methods generate immediate predictions and incrementally update the model upon detecting concept drift, such as using covariance matrices or least-square support vector machines. However, they often sacrifice performance for efficiency due to expensive computations. The MobileDA \cite{yang2020mobileda} addresses domain adaptation for edge devices by distilling knowledge from a teacher network on a server to a student network on the edge device, achieving domain-invariance and state-of-the-art performance in real-world scenarios like IoT-based WiFi gesture recognition. However, the effectiveness of MobileDA assumes that the teacher network has performed well on field data, which is challenging in practice because field data is often unlabelled and difficult to obtain. Relying exclusively on teacher models trained with web-sourced data can lead to deficiencies in the student model's capacity to collect data and self-improve. Suppose the knowledge embedded in the teacher model does not align with the student model’s operational environment. The student may face difficulties gathering and learning from relevant real-world data in that case. This misalignment can impede the student model's ability to adapt and evolve, compromising its effectiveness and overall utility.

Distributed and collaborative techniques are widely used for edge-device ML models. These techniques leverage the computational capabilities of multiple edge devices, aggregating their results instead of relying on a single resource-constrained device. Federated learning (FL) \cite{mcmahan2017communication,abreha2022federated, abbas2023novel, zhang2021federated} offers a transformative approach to decentralised model training. In the context of edge learning, where data is distributed across numerous edge devices, FL enables collaborative training without centralising sensitive data. While FL can help reduce computation needs for model training, addressing data shift and self-training for edge devices remains challenging.

\section{Methodology}
\subsection{Notation.} 
We denote our training set as \( \mathcal{D}_w \), consisting of
three components: field images \(\mathcal{I}_{\text{field}}\) with corresponding pseudo labels \(\mathcal{Y}_{\text{field}}\), synthesised images \(\mathcal{I}_{\text{syn}}\) with corresponding labels \(\mathcal{Y}_{\text{syn}}\), and web-sourced images \(\mathcal{I}_{\text{web}}\). The lightweight model deployed in the field is denoted as \(W_{\text{field}}\), while the cloud-based model is denoted as \(W_{\text{cloud}}\). Our objective is to train and continually improve \(W_{\text{field}}\) as data shifts occur without relying on additional manual annotations. This approach ensures that \(W_{\text{field}}\) remains effective and adaptive to new environmental conditions and variations.

\subsection{System Setup} 
\textbf{Sensors:} The system features a sensor suite comprising two RGB cameras and one thermal camera. The RGB cameras are Lucid Vision Labs TDR054S-CC with 5.4 MP resolution and 120 dB HDR imaging.
The cameras are IP67-rated (for operation in all weather conditions) and equipped with lenses of 41.4\textdegree{} horizontal field-of-view (HFoV) and 10\textdegree{} HFoV, to cover near and far scenes, respectively.
For nighttime detection, the thermal camera is a FLIR A68 with a 640 × 480 resolution and 24\textdegree{} HFoV housed in a weatherproof autoVimation Salamander enclosure.

\textbf{Edge Computer:} Networking and computing are managed by a QNAP QSW-2104-2T network switch and an NVIDIA Orin 64GB Dev Kit, respectively.

\textbf{ML Models:} Our system employs the
YOLOv8 \cite{yolov8} as \(W_{\text{field}} \), which is lightweight and computationally efficient for real-time object detection. We use LA-NMS with foundation models OWL-VIT \cite{minderer2024scaling} and Segment Anything Model (SAM) \cite{kirillov2023segment} as \( W_{\text{cloud}} \) for the auto-labelling of data.

\begin{figure*}[!t]
    \centering
    \includegraphics[width=0.9\textwidth]{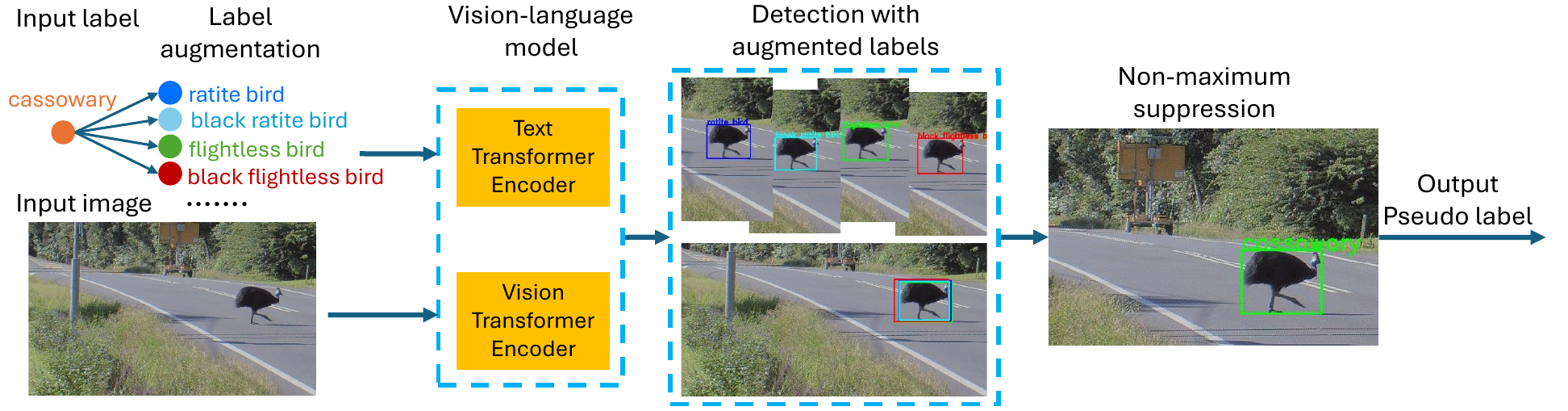}
    \caption{\small Workflow of the proposed LA-NMS. The approach begins by taking an input animal label and generating multiple related class labels. These augmented labels and input images are encoded using a text transformer encoder and a vision transformer encoder, respectively, within a VLM, i.e., OWL-VIT in this work. This process facilitates the identification of detection candidates in the input images based on the augmented labels. With detection results, the LA-NMS approach then applies NMS to eliminate redundant bounding boxes, selecting the most likely candidates as the output pseudo labels.}
    \label{fig:example-la-nms}
\end{figure*}

\subsection{Overview}

\textit{Fig. \ref{fig:stage_overview}} depicts an overview of the self-training pipeline for detecting rare animals. It is structured into two stages and integrates cloud and edge computing to create a robust and scalable system capable of adapting to different conditions.
The proposed process consists of two stages.
In \textit{Stage 1}, the proposed method uses \(W_{\text{cloud}}\) hosted on a cloud server to generate synthetic data \((\mathcal{I}_{\text{syn}}, \mathcal{Y}_{\text{syn}})\) for training the initial \(W_{\text{field}}\) for field deployment. This initial training set uses web-sourced data \(\mathcal{I}_{\text{web}}\).
In \textit{Stage 2}, the deployed model \(W_{\text{field}} \) processes camera images from the field and selects relevant information to be uploaded to the cloud server. \(W_{\text{cloud}}\) then automatically processes these field images \(\mathcal{I}_{\text{field}}\) to generate pseudo-labels \(\mathcal{Y}_{\text{field}}\) for fine-turning \(W_{\text{field}}\).
This dynamic updating mechanism in \textit{Stage 2} ensures the model evolves in response to data shifts.

\subsection{Stage 1: Synthetic Data Generation for Initialising Training Phase} 
Training ML models to accurately detect particular animal species typically requires a large amount of annotated high-quality data. However, there are challenges due to hardware limitations, data transfer constraints, and the underrepresentation of these animals in publicly available datasets. To address these challenges, we generated synthetic data \((\mathcal{I}_{\text{syn}}, \mathcal{Y}_{\text{syn}})\) based on \(\mathcal{I}_{\text{web}}\) using \(W_{\text{cloud}}\) in \textit{Stage 1}. This process involves using the proposed LA-NMS to expand the results of OWL-VIT \cite{minderer2024scaling}.

\subsubsection{Label-Augmentation Non-Maximum Suppression} \label{sec:la-nms}

Multi-modal image-text models like VLMs encode an image \(x\) and a class label \(c\) using an image encoder \(f_{\text{image}}\) and a text encoder \(f_{\text{text}}\), respectively. This encoding results in embeddings \(z_m = f_{\text{image}}(x)\) and \(z_c = f_{\text{text}}(c)\). The prediction logit score is calculated as the cosine similarity between these embeddings. The VLM then predicts the class that maximises this logit score:
\begin{equation}
\hat{c} = \arg \max_{c \in C} \text{logit}(x, c)
\end{equation}

To expand the search of rare species like cassowaries in images, we employ a label-augmentation strategy using the lexical database \textit{WordNet}~\cite{fellbaum2010wordnet}. For the target class \( c \), we identify parent and child classes \( H(c) \) that share visual similarities with \( c \) and augment the input text with these classes. For example, the class \emph{cassowary} can be augmented with its parent class \emph{flightless bird}. This set \(H(c)\) includes classes like the broader family or closely related species, enriching the label space and allowing the model to generalise better and differentiate between similar classes.

The augmented logit scores are computed by considering both the original and classes:
\begin{equation}
\text{logit}_{\text{aug}}(x, c) = \max(\text{logit}(x, c), \max_{h \in H(c)} \text{logit}(x, h))
\end{equation}

The prediction then adjusts to consider the highest-scoring class across both the direct and labels:
\begin{equation}
\hat{c}_{\text{aug}} = \arg \max_{c \in C} \text{logit}_{\text{aug}}(x, c)
\end{equation}

To manage the increased likelihood of multiple high-scoring but overlapping predictions due to the expanded label set, non-maximum suppression (NMS) is applied. NMS evaluates the confidence scores of these predictions and eliminates any that overlap, thus ensuring only the most relevant predictions are retained:
\begin{equation}
\text{\small NMS}(\{(c_i, s_i)\}) \rightarrow \{(c_j, s_j)\ |\ c_j \text{ are non-overlapping}\}
\end{equation}

\textit{Fig. \ref{fig:example-la-nms}} presents how LA-NMS is applied to an image containing a cassowary.

Subsequently, the output bounding boxes from the LA-NMS are used by the SAM \cite{kirillov2023segment} mask decoder to extract segmented instances of the animal, which are integrated into field background images using Gaussian blending  \cite{kirillov2023segment, ge2023beyond, dwibedi2017cut, ghiasi2021simple}.
At the end of \textit{Stage 1}, we train the initial model \(W_{\text{field}}\) with \(\mathcal{I}_{\text{syn}}\) and \(\mathcal{Y}_{\text{syn}}\) and deploy it in the field to gather real-world data of the animal species systematically.

\subsection{Stage 2: Self-Training for Field Data}

In \textit{Stage 2}, \(W_{\text{cloud}}\) processes the field images \(\mathcal{I}_{\text{field}}\) collected by \(W_{\text{field}}\) to generate pseudo-labels \(\mathcal{Y}_{\text{field}}\) using the proposed LA-NMS and refine the training dataset iteratively. Only a small subset (1.5\% in our field trial) of all recorded data are of interest for training and evaluation and thus are transferred from the edge computer to the cloud to conserve communication bandwidth. These uploaded data mainly include True Positive (TP) and False Positive (FP) detections by \(W_{\text{field}}\).
However, \(W_{\text{cloud}}\) often struggles to detect rare animal species accurately due to their underrepresentation in the training data. To mitigate this, we use the LA-NMS with a higher detection threshold to reduce the occurrence of mislabelling.
We then fine-tune \(W_{\text{field}}\) using the combination of the synthetic data \((\mathcal{I}_{\text{syn}}, \mathcal{Y}_{\text{syn}})\) and the newly generated auto-labelled field data \((\mathcal{I}_{\text{field}}, \mathcal{Y}_{\text{field}})\). The updated \(W_{\text{field}}\) is deployed in the field to collect further data for the next iteration of model fine-tuning. This iterative cycle progressively improves the detection performance of \(W_{\text{field}}\) in response to changes in the deployment environment.

\subsection{System Implementation}

\subsubsection{Data Logging} 
The data logging is divided into two parts: continuous data logging and event-triggered data logging. The continuous logging operates 24/7, recording field data used in \textit{Stages 1 and 2} at a lower frame rate. The event-triggered logging records images only for 10 to 15 seconds before and after detection; this information is used for post-analysis and event playback.

\subsubsection{Model Fine-Tuning and Domain Adaptation} 
A \(W_{\text{field}} \) pre-trained on the COCO dataset \cite{lin2014microsoft} was fine-tuned using synthetic data generated in \textit{Stage 1} and field data generated in \textit{Stage 2}. This fine-tuning process was conducted to adjust the model to the local conditions. 50\% of the last layers within the model \(W_{\text{field}}\) were trained while the remainder were frozen. To adapt \(W_{\text{field}}\) to the thermal image domain, we employ the same scheme as used for the RGB domain. We fine-tuned the COCO pre-trained model with synthetic thermal images generated in \textit{Stage 1} and real thermal images collected in \textit{Stage 2}.

\section{LA-NMS Validation}

To investigate the effectiveness of LA-NMS in enhancing the auto-labeling process, we evaluate the performance of the VLM with and without LA-NMS enabled on the public VOC dataset, focusing on 19 object classes. We exclude the \emph{human} class from our evaluation due to its high variability in appearances within images, which could introduce inconsistencies.
Our proposed LA-NMS integrates related classes provided by the VOC dataset. For example, the label augmentation for the \emph{sofa} class includes semantically related classes in \textit{WordNet}~\cite{fellbaum2010wordnet} such as \emph{studio couch, day bed, sofa, park bench}

\begin{figure}[b]
\centering
\includegraphics[width=\columnwidth]{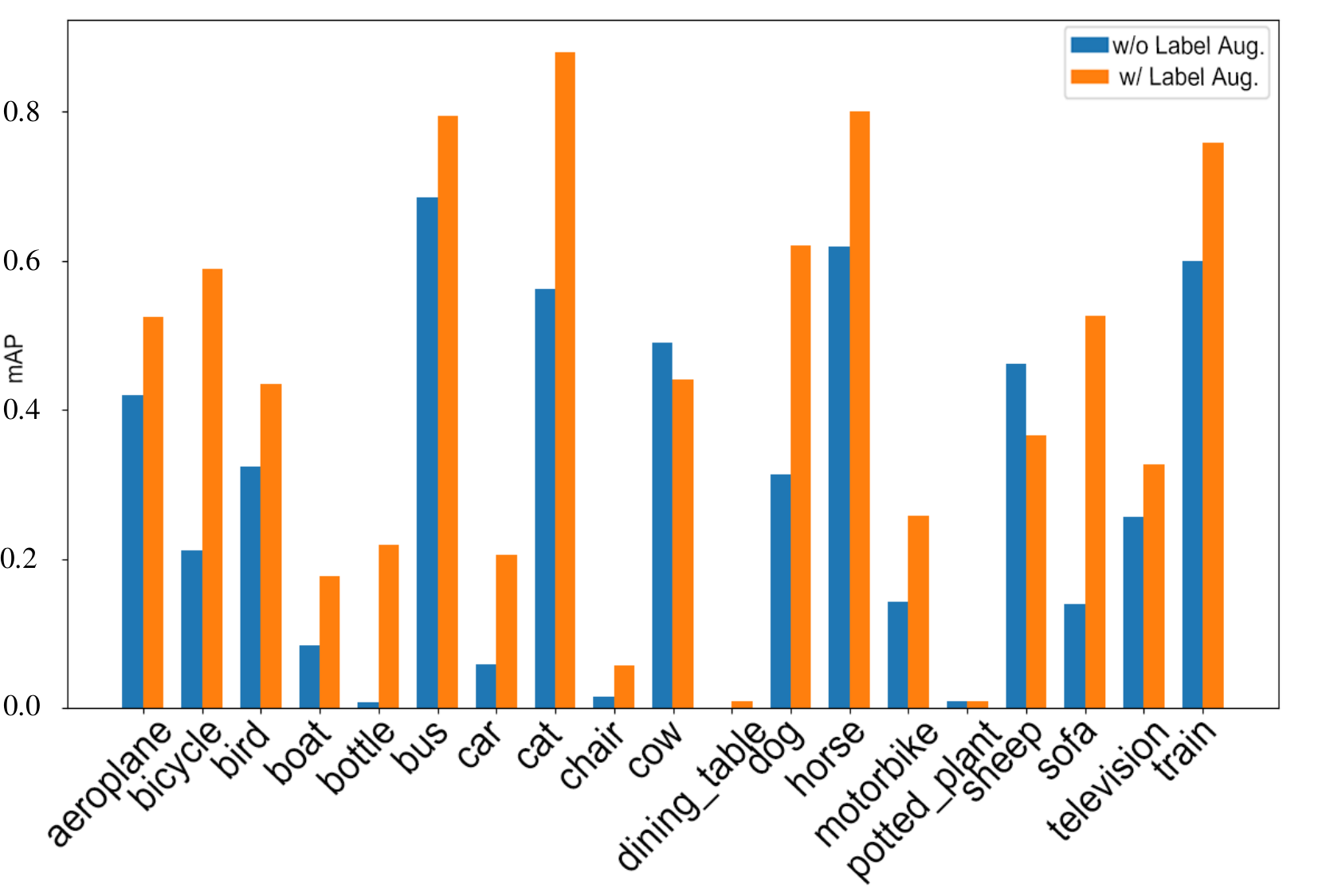}
\caption{\small Comparison of mAP for various object classes from the VOC dataset, analysed with and without LA-NMS. `Label Aug.' in the legend refers to LA-NMS.}
\label{fig:class_compare}
\end{figure}

We employ Mean Average Precision (mAP) calculated at an Intersection over Union (IoU) threshold of 0.5 as a metric that effectively balances precision and recall.  
\textit{Fig. \ref{fig:class_compare}} compares the mAP for 19 object classes from the VOC dataset. The model with LA-NMS performs better in 17 of these classes. For instance, the augmented labels for the class \emph{cat} include terms such as \emph{catamount}, \emph{cougar}, and \emph{lynx}, etc. Similarly, for the class \emph{bicycle}, the augmented labels include \emph{mountain bike}, \emph{all-terrain bike}, \emph{off-roader}, \emph{bicycle-built-for-two}, \emph{tandem bicycle}, \emph{tandem}, \emph{bicycle}, \emph{tricycle}, \emph{trike}, and \emph{velocipede}. The detection of both classes improves with the proposed LA-NMS method. However, the classes \emph{sheep} and \emph{cow} exhibit lower mAP. This decrease in performance is due to the augmented labels for these classes confusing the detection. For instance, the augmented labels for \emph{sheep} include \emph{llama}, \emph{ibex}, and \emph{cimarron}. The misclassification of some augmented classes caused an increase in FPs, leading to a drop in mAP.
It is therfore important to recognise that the environment and the augmented classes within that context influence LA-NMS's performance.

\section{Experiments}
This section details the experimental evaluations conducted using data from the field trial to assess various aspects of our proposed pipeline.

\subsection{Datasets}
The field-collected data is divided into training and evaluation sets. The training data were used to train the cassowary detection models. There are, in total, 10 models (1 in \textit{Stage 1} and 9 in \textit{Stage 2}) trained during the data collection and on-road trial periods, using data available up to different dates. Most of the model training work occurred during the data collection period, i.e., from February to April 2024, in preparation for the subsequent on-road trial in May and June 2024. The following datasets are used for training and evaluating the proposed method:

\subsubsection{Training dataset}
\begin{itemize}
        \item Field-collected subset: We train the models with raw field data with pseudo-labels. This dynamically updated subset is maintained at 4,000 to 6,000 images each time, including cassowary and non-cassowary images. The validation subset is also dynamically updated. It contains approximately 1,000 images with cassowaries.
        \item Web-sourced subset: Consists of 89 object-centric RGB images of cassowaries sourced from the internet and 7 thermal images from \cite{eastick2019cassowary}.
\end{itemize}

\subsubsection{Evaluation Dataset}
\label{sec:evaluation_dataset}
\begin{itemize}    
        \item Field-collected cassowary subset: This subset contains 4577 labelled images (3127 RGB + 1450 thermal images), containing cassowaries from all cameras over 38 cassowary sighting cases in the field.
        \item Field-collected non-cassowary subset: Contains 11975 images (9166 RGB + 2809 thermal) without cassowaries from all cameras.
    \end{itemize}

\subsection{Evaluation Metrics}
\begin{itemize}
\item \textit{Mean True Positive Rate (mTPR)} quantifies the model’s ability to identify the target objects within the dataset.
For each image in the test dataset containing cassowaries, we compute the True Positive Rate (TPR). The mTPR is then obtained by averaging these TPR values.
A higher mTPR indicates better detection performance.
\item \textit{False Positive Rate (FPR)} evaluates the frequency with which the model incorrectly detects a cassowary, providing insight into the potential for generating false alarms. The FPR is calculated based on all RGB images in the evaluation dataset's non-cassowary subset.
\end{itemize}

\subsection{Methodology}
To validate the method in a real-world application, we apply the proposed approach to the \textit{cassowary} in the field data, using labels like \emph{flightless bird, ratite, ratite bird}. To enrich the label augmentation, we incorporate visual descriptors, such as colour, resulting in augmented labels like \emph{black flightless bird, black ratite, black ratite bird}.

\subsubsection{Model Self-Improvement in Field Environment:} To validate the proposed method for ongoing enhancement during various stages of real-world deployment, we train \(W_{\text{field}}\) with data from \textit{Stage 1} and \textit{Stage 2} on the field-collected data. Additionally, we assess the proposed method through field trials by comparing models trained during data collection and on-road trial periods using data available up to different dates. This implies that the models were trained and validated with distinct datasets. We evaluate each model on the field evaluation dataset to ensure a fair comparison between each model, as presented in \textit{Sec. \ref{sec:evaluation_dataset}}.

\begin{figure}[b!]
\centering
\includegraphics[width=\columnwidth]{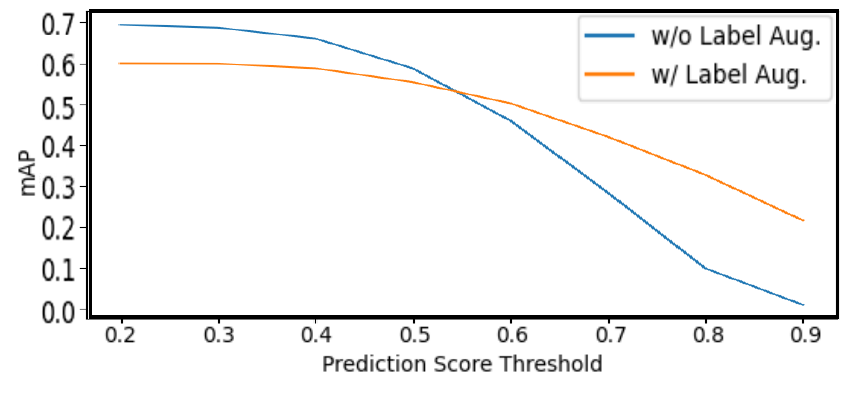}
\caption{\small Performance of the VLM with and without LA-NMS enabled across various threshold settings. `Label Aug.' in the legend refers to LA-NMS. The results were obtained based on the field evaluation dataset.}
\label{fig:map_labaug_compare}
\end{figure}

\begin{figure*}[!t]
	\centering
	\begin{subfigure}[]{5.8cm}
        \label{fig:roc:a}
        \includegraphics[trim={1.2cm 0.3cm 1.6cm 1.4cm},clip, width=5.5cm]{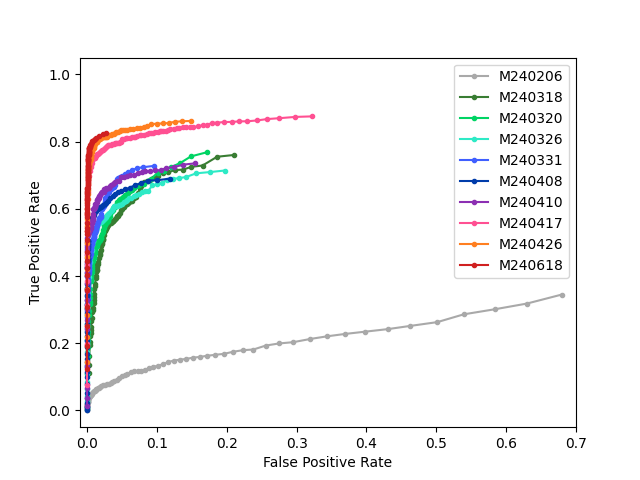}
        \caption{ROC curve at 0--100\,m}
    \end{subfigure}
    \begin{subfigure}[]{5.8cm}
        \label{fig:roc:b}
        \includegraphics[trim={1.2cm 0.3cm 1.6cm 1.4cm},clip,width=5.5cm]{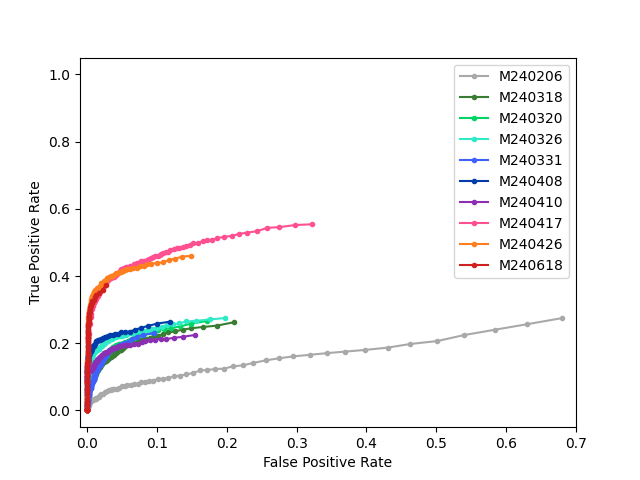}
        \caption{ROC curve at 100--200\,m}
    \end{subfigure}
    \begin{subfigure}[]{5.8cm}
        \label{fig:roc:c}
        \includegraphics[trim={1.2cm 0.3cm 1.6cm 1.4cm},clip,width=5.5cm]{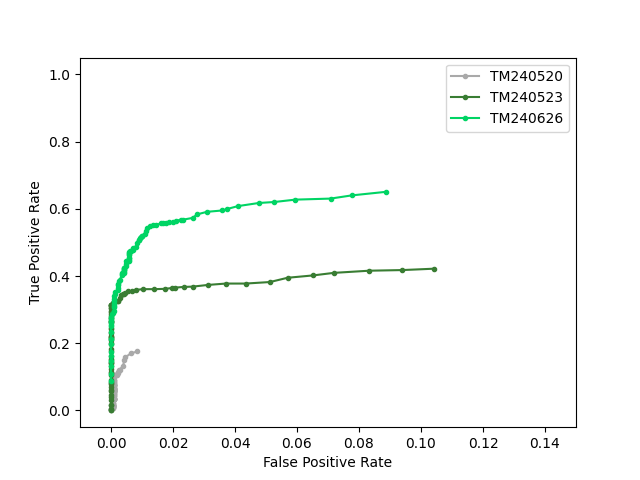}
        \caption{ROC curve for thermal images at 0--100\,m}
    \end{subfigure}
    \caption{\small ROC curves of the trained models on RGB and thermal images for cassowary crossing scenes.
    }
	\label{fig:roc}
\end{figure*}

\subsubsection{Adaptability to Thermal Domain:} The proposed method's adaptability is tested to assess its capability from the RGB domain to the thermal domain. This experiment is also crucial for validating its robustness and effectiveness in environments where thermal imaging is necessary due to challenging lighting conditions unsuitable for RGB sensors.

\section{Results and Discussions}
\subsection{Evaluation of LA-NMS in Auto-Labelling Enhancements}
This section presents the performance improvements achieved through the proposed LA-NMS on the field data.

\textit{Fig. \ref{fig:map_labaug_compare}} illustrates the performance comparison between the VLM with and without LA-NMS enabled across various confidence threshold settings on the field evaluation dataset. Calibration of the confidence threshold is critical in practical applications to ensure pseudo-labels' accuracy and minimise FPs. A lower threshold often results in a higher incidence of FPs. In our self-training pipeline, we adopt a higher threshold to filter out misdetections. The experimental results demonstrate that LA-NMS enhances model performance, particularly when the confidence threshold exceeds 0.5. As the threshold increases, the model equipped with LA-NMS consistently outperforms the baseline model without LA-NMS, achieving better mAP. This improvement is attributed to the label augmentation process, which enables the detection model to consider a broader range of potential candidates and refine target detections through NMS. 

\textit{Tab. \ref{tab:label_aug_comp}} also shows that LA-NMS significantly improves the VLM's performance in detecting cassowaries in both field data \(\mathcal{I}_{\text{field}}\) and web-sourced data \(\mathcal{I}_{\text{web}}\).

\begin{table}[btp]
\centering
\begin{tabular}{ccc}
\toprule
Method         & Field Data & Web-Sourced Dataset \\
\midrule
w/o LA-NMS & 0.683          & 0.472                \\
w/ LA-NMS  & 0.795          & 0.529          \\
\bottomrule
\end{tabular}
\caption{\small mAP of the VLM with and without LA-NMS enabled for field data and web-sourced data.}
\label{tab:label_aug_comp}
\end{table}

\subsection{Model Self-Improvement in Field Environment}

\begin{table}[]
\centering
\begin{tabular}{ccccc}
\toprule
Stage & Model Name & \makecell{mTPR\\(0--100\,m)} & \makecell{mTPR\\(100--200\,m)} & FPR\\
\midrule
1 & M240206 & 4.2\%  & 2.6\% & 0.37\% \\
\midrule
\multirow{9}{*}{2} & M240318 & 19.4\%  & 10.6\%  & 0.39\% \\
& M240320 & 32.3\%  & 6.9\%  & 0.37\% \\
& M240326 & 45.3\%  & 14.7\%  & 0.38\% \\
& M240331 & 45.8\%  & 6.4\%  & 0.39\% \\
& M240408 & 54.5\%  & 17.1\%  & 0.37\% \\
& M240410 & 53.6\%  & 11.8\%  & 0.38\% \\
& M240417 & 71.3\%  & 25.7\%  & 0.37\% \\
& M240426 & 73.7\%  & 29.9\%  & 0.35\% \\
& M240618 & 78.5\%  & 30.0\%  & 0.37\% \\
\bottomrule
\end{tabular}
\caption{\small The evaluation results of 10 trained \(W_{\text{field}}\) for cassowary detection in the field trial. The models are listed in chronological order of their training dates (M year/month/date). For each model, the TPR results are first calculated for RGB frames in every sighting case in the cassowary subset of the field evaluation dataset and then averaged over all cases in each range group. The FPR results are calculated based on all RGB frames combined in the non-cassowary subset of the dataset.}
\label{tab:model_vs_range}
\end{table}

This section presents the field evaluation results of the proposed self-training scheme. To fairly evaluate the 10 trained \(W_{\text{field}}\) during the field trial, an FPR cut-off threshold of 0.4\% is employed in the model evaluation, with the results presented in \textit{Tab. \ref{tab:model_vs_range}}. The results demonstrate a significant performance improvement from \textit{Stage 1} to \textit{Stage 2}. The results also showcase a clear trend that models trained later achieve higher mTPR in \textit{Stage 2}, primarily due to the incorporation of more field data in training. Specifically, the initial model, M240206, was trained with synthetic data only in \textit{Stage 1}. It starts with a low mTPR of 4.2\% for detections up to 100\,m and 2.6\% for detections within the range of 100--200\,m. The progressive performance improvement in \textit{Stage 2} is evident as the model underwent training with additional real-world data captured from the field. The best-performing models are M240426 and M240618. The model M240426, which was used throughout the on-road trial period from May to June, achieves mTPR of 73.7\% and 29.9\% for the 0--100\,m and 100--200\,m ranges, respectively. Compared with earlier models, M240628 was trained with the largest dataset from March to early June. It achieves the highest mTPR, i.e., 78.5\% and 30.0\%, for the two ranges, respectively. Although this model was not deployed in the field as it was close to the end of the on-road trial, the results significantly validate the proposed self-training scheme.

\begin{figure*}[t]
\centering
\includegraphics[width=\textwidth]{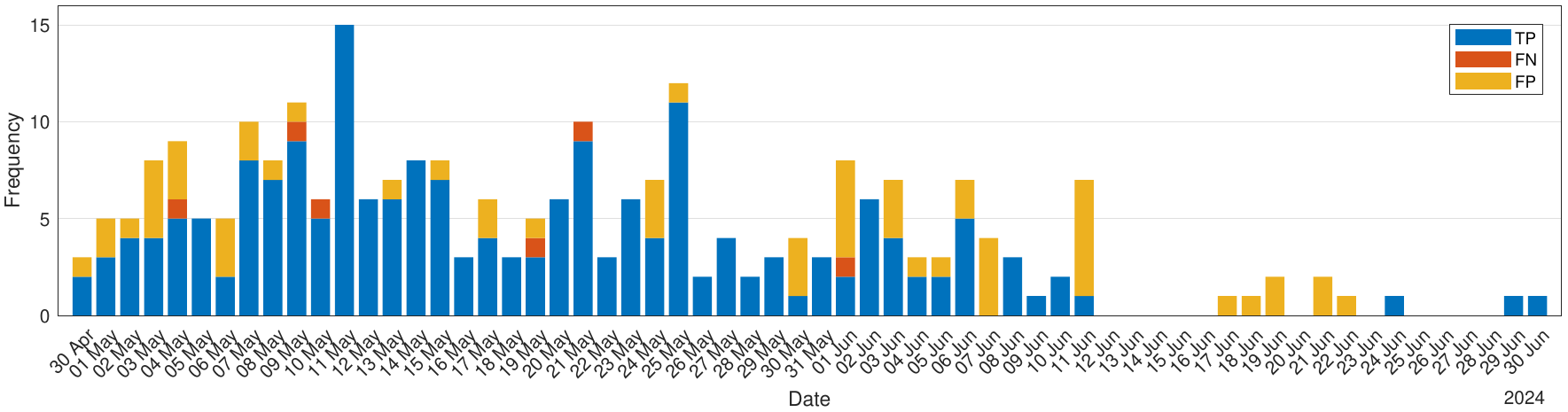}
\caption{\small Distribution of cassowary detection events over time.
There were 166 cassowary sighting cases in May 2024 before a steep drop in June 2024. Each event has been checked against the ground truth data to classify them as TP, FN, and FP detection events. Despite the system running, there were no cassowary activities for up to two weeks in June, specifically from 12 June to 23 June 2024.
}
\label{fig:cases_vs_date}
\end{figure*}

In addition, a collection of Receiver Operating Characteristic (ROC) curves are presented in \textit{Fig. \ref{fig:roc}(a)} and \textit{Fig. \ref{fig:roc}(b)} to summarise the detection performance for different models and ranges. It is seen that the overall performance of the models M240618, M240426, and M240417 are close and are significantly better than that of others.

The model evaluation results above do not fully reflect the overall cassowary detection system's performance. The system's detection events are initiated by an event-triggering pipeline running on the edge computer, which aggregates the Bayesian-filtered results from multiple instances of \(W_{\text{field}}\) across different cameras. The performance of the detection system based on model M240426 is evaluated based on detection events recorded during the on-road trial, which lasted from 30 April to 30 June 2024. During the on-road trial, the system triggered VMS alerts as presented in \textit{Fig. \ref{fig:system}(b)} for motorists upon detecting a cassowary on or near the road. Cassowary sightings during the trial period were recorded as the ground truth data. The detection models first reported the sightings, which were manually verified in the recorded videos. It should be emphasised that manual data inspection in this work is solely for performance evaluation. No manual data labelling is required for the developed self-training scheme.

Over the 62 days of the on-road trial period, the detection system reported 259 events. Each event has been checked against the ground truth data to classify them as TP, False Negative (FN), and FP detection events. The distribution of different detection events over the trial dates is presented in \textit{Fig. \ref{fig:cases_vs_date}}. Each event type's total and average numbers are summarised in \textit{Tab. \ref{tab:events_summary}}, which highlights that the system missed as few as six cassowary cases during the trial period. These FN events were identified through manual inspection of the recorded data.

Based on the summarised events, the system's precision and recall during the on-road trial are 0.77 and 0.97, respectively. The exceptionally high recall demonstrates the system's high sensitivity in detecting cassowaries crossing the road or on the roadsides, a critical aspect for road safety-related use cases, as \textit{Tab. \ref{tab:events_summary}} also presents the system reported, on average, less than one FP event per day. This average FP result is reasonable, considering the deployed detection model examines 1.4 million RGB images daily from 6 am to 7 pm. The causes of the FP events primarily fall into a few categories: vehicles, persons, vegetation, animals, and shadows. Note that many FP events were caused by the same objects over a short period.

\begin{table}
    \centering
    \begin{tabular}{cccc}
        \toprule
        & TP & FN & FP \\
        \midrule
        Total & 194  & 6 & 59 \\
        Average (per day) & 3.13 & 0.10 & 0.95 \\
        \bottomrule
    \end{tabular}
    \caption{\small A summary of detection events during the 62-day on-road trial.}
    \label{tab:events_summary}
\end{table}

It is also shown in \textit{Fig. \ref{fig:cases_vs_date}} that FP events increased slightly in early June 2024. This is primarily because the model M240426 was trained with data collected in March and April but was deployed in the field from May to June. The weather in May was similar to that in earlier months, but the weather changes in June caused data shifts and, thus, more FP events in cassowary detection.

Despite these challenges, applying continual learning from field data enabled the self-training method to effectively reduce the FP rate by incorporating similar FP cases into model training. For instance, the model M240618 was trained using a dataset containing more recent data than the model used during the on-road trial period and has shown improved overall performance, as shown in \textit{Tab. \ref{tab:model_vs_range}}. Overall, the trend of performance improvement in \textit{Tab. \ref{tab:model_vs_range}} highlights the effectiveness of the self-training process, which iteratively refines the model's accuracy and reliability by learning from real-world data, ultimately achieving an optimal balance between detecting cassowaries and minimising false alerts.

\subsection{Thermal Domain Adaptation}

\begin{figure}[h]
\centering
    \begin{subfigure}[]{0.48\columnwidth}
        \label{fig:cass_syn1:a}
        \includegraphics[width=\columnwidth]{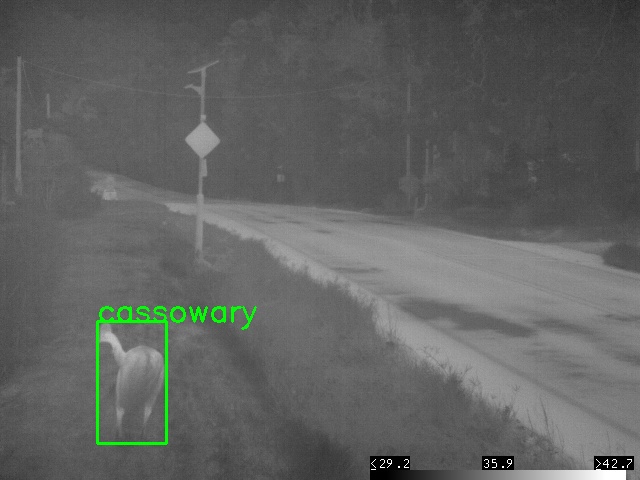}

    \end{subfigure}
    \begin{subfigure}[]{0.48\columnwidth}
        \label{fig:cass_syn1:b}
        \includegraphics[width=\columnwidth]{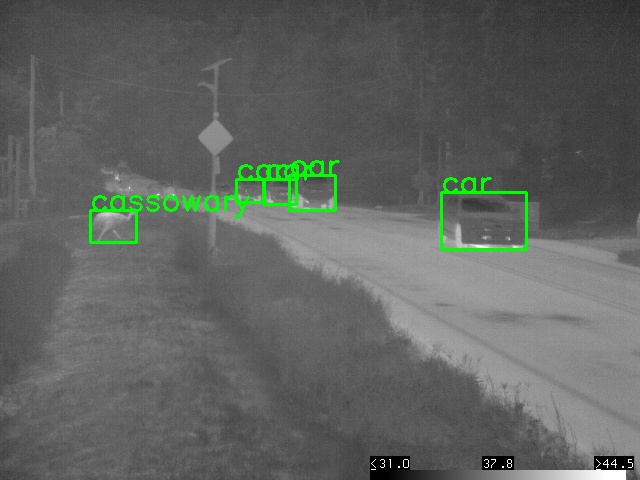}

    \end{subfigure}
\caption{\small Demonstrating the system's detection of a cassowary in thermal imagery from the field trial.}
\label{fig:cass_syn1}
\end{figure}

The experiment evaluates the adaptability of the proposed method to the thermal domain, which is essential for validating its robustness under inadequate lighting conditions.
\textit{Fig. \ref{fig:roc}(c)} demonstrates that the method consistently improves in accuracy and adaptability within challenging thermal imaging contexts in the field. Two examples of detecting cassowaries in thermal imaging during the field trial are presented in \textit{Fig. \ref{fig:cass_syn1}}. These findings substantiate the method’s capacity for adaptation across different domains and its continuous improvement through iterative learning, ensuring reliable performance across diverse operational environments.

\section{Conclusions}
In conclusion, this work has demonstrated the effectiveness of our self-training scheme for detecting rare animals on and near roads. By leveraging LA-NMS, the proposed method has shown improvements in the accuracy of animal detection. The robustness and adaptability of the proposed scheme were validated through experiments and a five-month field deployment. This research emphasises the effective deployment of automatic systems in resource-constrained environments to mitigate risks related to road safety and wildlife preservation, thus paving the way for expanded applications of such approaches.

{\small
\bibliographystyle{IEEEtran}
\bibliography{egbib}
}

\end{document}